# Explainable Artificial Intelligent (XAI) for Predicting Asphalt Concrete Stiffness and Rutting Resistance: Integrating Bailey's Aggregate Gradation Method

Warat Kongkitkul, Sompote Youwai [1], Siwipa Khamsoy and Manaswee Feungfung

AI research Group
Department of Civil Engineering
Faculty of Engineering
King Mongkut's University of Technology Thonburi

[1]Corresponding author

## Abstract

This study employs explainable artificial intelligence (XAI) techniques to analyze the behavior of asphalt concrete with varying aggregate gradations, focusing on resilience modulus (MR) and dynamic stability (DS) as measured by wheel track tests. The research utilizes a deep learning model with a multi-layer perceptron architecture to predict MR and DS based on aggregate gradation parameters derived from Bailey's Method, including coarse aggregate ratio (CA), fine aggregate coarse ratio (FAc), and other mix design variables. The model's performance was validated using k-fold cross-validation, demonstrating superior accuracy compared to alternative machine learning approaches. SHAP (SHapley Additive exPlanations) values were applied to interpret the model's predictions, providing insights into the relative importance and impact of different gradation characteristics on asphalt concrete performance. Key findings include the identification of critical aggregate size thresholds, particularly the 0.6 mm sieve size, which significantly influences both MR and DS. The study revealed size-dependent performance of aggregates, with coarse aggregates primarily affecting rutting resistance and medium-fine aggregates influencing stiffness. The research also highlighted the importance of aggregate lithology in determining rutting resistance. To facilitate practical application, web-based interfaces were developed for predicting MR and DS, incorporating explainable features to enhance transparency and interpretation of results. This research contributes a data-driven approach to understanding the complex relationships between aggregate gradation and asphalt concrete performance, potentially informing more efficient and performance-oriented mix design processes in the future.

## 1. Introduction

The performance and longevity of asphalt pavements, critical components of modern transportation infrastructure, are intrinsically linked to the complex interplay of their constituent materials. Aggregate gradation, in particular, plays a pivotal role in determining key properties such as durability, stability, and resistance to deformation (Garcia *et al.* 2020, Khasawneh *et al.* 2022). Despite extensive research, the intricate relationships between aggregate gradation and asphalt concrete behavior continue to challenge engineers and researchers in the field of intelligent transportation systems. Current practices in asphalt mixture design rely heavily on empirical methods and experience-based decisions, which often fall short in optimizing mixture performance across diverse environmental conditions and traffic loads(Lee *et al.* 2023, Zhang *et al.* 2023).

The complexity of interactions between aggregate gradation, binder properties, and external factors makes it challenging to predict pavement performance accurately, leading to suboptimal designs, premature failures, and increased lifecycle costs of transportation infrastructure (Khasawneh *et al.* 2022). While numerous studies have investigated the effects of aggregate gradation on asphalt concrete properties, there remains a significant gap in understanding the complex, non-linear relationships between gradation parameters and pavement performance metrics. Recent advancements in artificial intelligence, particularly in the field of explainable AI (XAI), present unprecedented opportunities to elucidate these complex patterns and relationships. XAI techniques provide interpretable insights into the decision-making processes of AI models, bridging the gap between black-box predictions and actionable engineering knowledge (Chaddad *et al.* 2023)

The motivation for using XAI techniques in this study is multifaceted. Firstly, XAI enhances interpretability, allowing for a deeper understanding of how different gradation parameters influence asphalt concrete performance (Ding and Kwon 2024). Secondly, it improves trust and adoption of AI-driven design processes by making predictions more transparent and explainable. Thirdly, XAI techniques can uncover complex, non-linear relationships between input variables and performance metrics that may not be apparent through traditional analysis methods (Hsiao *et al.* 2024). Lastly, the insights provided by XAI can guide targeted improvements in mixture design, potentially leading to more efficient and performance-oriented optimization processes (Abdollahi *et al.* 2024).

This study aims to address these research gaps by integrating the Bailey's method for aggregate gradation analysis (Bailey and Burch 2002) with SHAP (SHapley Additive exPlanations) values, a powerful XAI technique (Li *et al.* 2024). By focusing on two critical performance indicators—resilience modulus and resistance to rutting as measured by the wheel track test—this research seeks to elucidate the intricate relationships between gradation parameters and asphalt concrete performance within the context of intelligent transportation systems. The findings from this study have the potential to refine mix design practices, optimize pavement performance, and contribute to the development of more durable, cost-effective, and intelligent transportation infrastructure. By addressing the current limitations in asphalt mixture design and leveraging advanced AI techniques, this research aims to bridge the gap between traditional engineering practices and cutting-edge data science, paving the way for the next generation of intelligent transportation systems.

This study comprises two primary components: experimental investigation and computational modeling. The experimental phase focused on characterizing the mechanical behavior of asphaltic concrete, specifically its stiffness and rutting resistance, under varying aggregate gradations. Two lithologies, limestone and basalt, were employed to assess the impact of rock type on material properties. Stiffness was quantified via resilience modulus testing, adhering to ASTM D4123 (ASTM International 1995) and AASHTO TP 31 protocols (American Association of State Highway and Transportation Officials (AASHTO) 1994). Rutting resistance was evaluated using the Hamburg wheel-tracking (HWT) test, following AASHTO T324 (American Association of State Highway and Transportation Officials (AASHTO) 2022). standards. The computational phase involved the development of a predictive model based on a multilayer perceptron (MLP) architecture. This model was subsequently interpreted using an explainable AI (XAI) framework to elucidate the relationships between input parameters and predicted outcomes. To facilitate practical application, a web-based interface was implemented, integrating the trained model with XAI functionalities. This interface enables civil engineers to utilize advanced AI techniques without requiring in-depth knowledge of machine learning, while also providing interpretable insights into predictions and allowing for iterative optimization of aggregate gradations to achieve desired asphaltic concrete properties. This approach represents a novel integration of AI techniques with domain-specific engineering knowledge, potentially enhancing the efficiency and effectiveness of asphaltic concrete design processes.

## 2 Background

The Bailey's method (Bailey and Burch 2002), developed by Robert Bailey of the Illinois Department of Transportation in the 1980s, is a systematic approach to designing and analyzing aggregate gradations in asphalt mixtures. It focuses on understanding and controlling the packing characteristics of aggregates, using specific control sieves to divide the aggregate blend into coarse and fine fractions. The method employs three key ratios - Coarse Aggregate Ratio (CA Ratio), Fine Aggregate Coarse Ratio (FAc Ratio), and Fine Aggregate Fine Ratio (FAf Ratio) - to evaluate and adjust aggregate gradations. By controlling these ratios within these ranges, the Bailey's method aims to achieve desired volumetric properties and optimize mixture performance, potentially improving properties like rutting resistance and durability. This approach provides engineers with a tool to better understand and control the aggregate structure in asphalt mixtures, offering flexibility to accommodate various aggregate types and mixture requirements. In the context of the research paper discussed, these Bailey's method parameters were used as inputs for the deep learning model, enabling a comprehensive analysis of how gradation characteristics influence asphalt concrete performance metrics such as resilience modulus and dynamic stability.

According to the Bailey's method for the design of coarse-grade mixes, various particle sizes are defined using specific control sieve sizes as follows: The Half Sieve (HS) is set at 0.5 times the Nominal Maximum Particle Size (NMPS). The Primary Control Sieve (PCS), which separates coarse aggregate from fine aggregate (Figure 1), is defined as PCS = 0.22 × NMPS. The Secondary Control Sieve (SCS) separates the coarse and fine portions of the fine aggregate and is defined as SCS = 0.22 × PCS. Additionally, the fine portion of the fine aggregate is further separated using the Tertiary Control Sieve (TCS), defined as TCS = 0.22 × SCS.

The Coarse Aggregate Ratio (CA Ratio) evaluates the packing of coarse aggregates and resulting void structure (Figure 1). Particles retained on the half sieve ("interceptors") and those passing it ("pluggers") interact to influence the mixture's volumetric properties. Adjusting the balance of these particles can modify the Voids in Mineral Aggregate (VMA), potentially optimizing compactability and load performance (Vavrik *et al.* 2002). The CA Ratio calculation is expressed as follows:

$$CA = \frac{\% \text{ Passing Half Sieve} + \% \text{ Passing } PCS}{100\% - \% \text{ Passing Half Sieve}} \qquad (1)$$

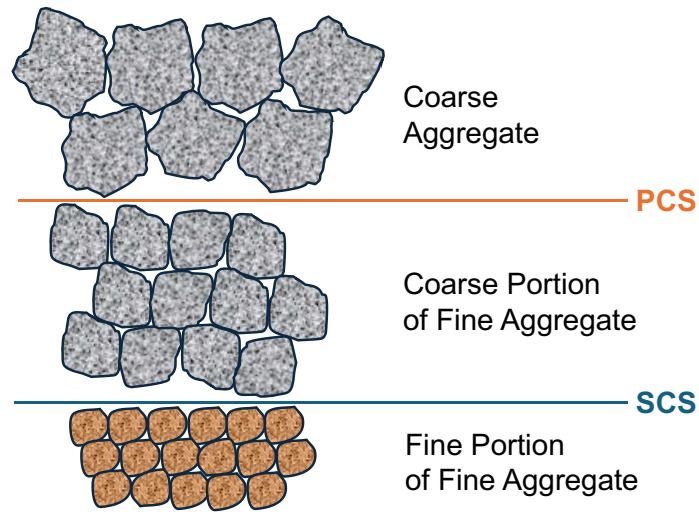

**Figure 1** Aggregate Divisions in a Continuous Gradation Based on the Bailey's method.

The Fine Aggregate Coarse Ratio (*FAc* Ratio) examines the aggregate material passing the PCS and retained on the SCS. It evaluates the coarse portion of the fine aggregate, which creates voids to be filled with the fine portion. Proper filling of these voids controls the Voids in Mineral Aggregate (VMA) and air voids in the mixture. Equation 2 expresses the mathematical formulation of the *FAc* Ratio.

$$FAc = \frac{\%\text{Passing SCS}}{\%\text{Passing PCS}} \qquad (2)$$

The Fine Aggregate Fine Ratio (*FAf* Ratio) is similar to the *FAc* Ratio but uses the TCS to evaluate the fine portion of the fine aggregate. It fills the voids created by the coarse portion of the fine aggregate. For dense-graded mixtures, the *FAf* Ratio should be less than 0.50, with VMA increasing as the ratio decreases. The equation for the *FAf* Ratio is provided in Eq. 2.

$$FAf = \frac{\%\text{Passimp TCS}}{\%\text{Passimp SCS}} \qquad (3)$$

## 3. Laboratory testing
### 3.1 Aggregate gradation

This study used two types of aggregate: limestone and basalt, each supplied from four bins with varying gradations. The final gradation was achieved by blending these bins in different proportions. The NMPS for all bins is 19 mm. Bailey's method, originally developed for coarse-graded mixes, recommends the following ratios for a 19 mm NMPS: a CA Ratio of 0.60–0.75, an *FAc* Ratio of 0.35–0.50, and an *FAf* Ratio of 0.35–0.50 (Vavrik *et al.* 2002). This study prepared various gradations based on these recommended values by combining the lower, mid-range, and upper bound values for the CA, *FAc*, and *FAf* ratios, resulting in 3 × 3 = 9 combinations, as shown in Table 1. Each gradation is named with the second and fourth letters indicating the lower (L), mid-range (M), or upper (U) bound values of the coarse (C) or fine (F) portions, specified by the first and third letters, respectively.

**Table 1.** Bailey Ratios for Different Aggregate Gradations, AC Contents for AV = 4%, and Measured $M_r$ Values from Resilient Modulus Tests and DS Values from HWT Tests

| No. | Gradation's name | CA[+] | FAc[+] | FAf[+] | new CA[$] | new FAc[$] | new FAf[$] | Limestone | | | Basalt | | |
|---|---|---|---|---|---|---|---|---|---|---|---|---|---|
| | | | | | | | | AC content to achieve AV = 4%[#] | Average $M_R$[&] (MPa) | Average DS[%] (pass/mm) | AC content to achieve AV = 4%[#] | Average $M_R$[&] (MPa) | Average DS[%] (pass/mm) |
| 1 | CLFL | 0.600 | 0.350 | 0.350 | 0.600 | 0.350 | 0.350 | 4.82 | 1167.9 | 369 | 6.02 | 1144.5 | 338 |
| 2 | CLFM | 0.600 | 0.425 | 0.425 | 0.600 | 0.425 | 0.425 | 4.29 | 1329.0 | 468 | 5.91 | 1286.5 | 265 |
| 3 | CLFU | 0.600 | 0.500 | 0.500 | 0.600 | 0.500 | 0.500 | 4.23 | 1375.2 | 506 | 5.76 | 1323.0 | 249 |
| 4 | CMFL | 0.675 | 0.350 | 0.350 | 0.800 | 0.350 | 0.350 | 4.98 | 1189.8 | 452 | 6.07 | 1241.8 | 513 |
| 5 | CMFM | 0.675 | 0.425 | 0.425 | 0.800 | 0.425 | 0.425 | 4.36 | 1418.4 | 554 | 5.79 | 1335.0 | 348 |
| 6 | CMFU | 0.675 | 0.500 | 0.500 | 0.800 | 0.500 | 0.500 | 3.94 | 1662.9 | 603 | 5.57 | 1366.3 | 307 |
| 7 | CUFL | 0.750 | 0.350 | 0.350 | 1.000 | 0.350 | 0.350 | 4.82 | 1230.1 | 498 | 5.82 | 1258.3 | 566 |
| 8 | CUFM | 0.750 | 0.425 | 0.425 | 1.000 | 0.425 | 0.425 | 4.56 | 1504.5 | 620 | 5.50 | 1345.4 | 487 |
| 9 | CUFU | 0.750 | 0.500 | 0.500 | 1.000 | 0.500 | 0.500 | 4.14 | 1725.3 | 755 | 5.47 | 1419.2 | 335 |

[+]: for coarse-graded mixes; [$]: for fine-graded mixes; [#]: determined by Marshall's method; [&]: resilient modulus; and [%]: dynamic stability from Hamburg wheel-tracking test

However, the aggregate gradation commonly used by the Department of Highways in Thailand, where this study was conducted, is fine-graded. Therefore, the designed gradation must satisfy not only the CA, FAc, and FAf ratios for coarse-graded mixes but also the new ratios for fine-graded mixes to properly control the fine portion. The new control sieve sizes are defined as follows: the new NMPS is the original PCS; the new HS is 0.5 × new NMPS; the new PCS is 0.22 × new NMPS; the new SCS is 0.22 × new PCS; and the TCS is 0.22 × new SCS. Table 3.6 also lists the new CA, FAc, and FAf ratios for fine-graded mixes based on the specified ratios for coarse-graded mixes.

In this study, the NMPS is 19 mm (3/4" sieve). For coarse-graded mixes, the HS is 9.5 mm (3/8" sieve), the PCS is 4.75 mm (#4 sieve), the SCS is 1.18 mm (#16 sieve), and the TCS is 0.3 mm (#50 sieve). For fine-graded mixes, the new NMPS is 4.75 mm (#4 sieve), the new HS is 2.36 mm (#8 sieve), the new PCS is 1.18 mm (#16 sieve), the new SCS is 0.30 mm (#50 sieve), and the new TCS is 0.075 mm (#200 sieve). Table 2 summarizes these original and new control sieve sizes for both coarse- and fine-graded mixes, along with the percent finer for each sieve size across the nine specified gradation characteristics.

**Table 2** Percent Finer for Specified Sieve Sizes for the Nine Gradation Characteristics Defined by the Bailey's Method Used in This Study

| No. | Sieve Size | | Coarse-graded mix | Fine-graded mix | Percent Finer | | | | | | | | |
|-----|------|------|--------|--------|-------|-------|-------|-------|-------|-------|-------|-------|-------|
| | in. | mm | | | CLFL | CLFM | CLFU | CMFL | CMFM | CMFU | CUFL | CUFM | CUFU |
| 1 | 3/4" | 19 | NMPS | | 100 | 100 | 100 | 100 | 100 | 100 | 100 | 100 | 100 |
| 2 | 1/2" | 12.5 | | | 86.10 | 86.10 | 86.10 | 87.45 | 87.45 | 87.45 | 88.35 | 88.35 | 88.35 |
| 3 | 3/8" | 9.5 | HS | | 69.82 | 69.82 | 69.82 | 71.32 | 71.32 | 71.32 | 72.76 | 72.76 | 72.76 |
| 4 | #4 | 4.75 | PCS | NMPS | 51.66 | 51.66 | 51.66 | 51.95 | 51.95 | 51.95 | 52.27 | 52.27 | 52.27 |
| 5 | #8 | 2.36 | | HS | 30.67 | 33.10 | 35.52 | 33.19 | 35.36 | 37.52 | 35.28 | 37.24 | 39.20 |
| 6 | #16 | 1.18 | SCS | PCS | 18.08 | 21.96 | 25.83 | 18.18 | 22.08 | 25.98 | 18.29 | 22.21 | 26.13 |
| 7 | #30 | 0.6 | | | 14.40 | 14.40 | 14.40 | 14.34 | 14.34 | 14.34 | 14.28 | 14.28 | 14.28 |
| 8 | #50 | 0.3 | TCS | SCS | 6.33 | 9.33 | 12.92 | 6.36 | 9.38 | 12.99 | 6.40 | 9.44 | 13.07 |
| 9 | #100 | 0.15 | | | 3.72 | 4.50 | 7.20 | 3.67 | 4.43 | 7.07 | 3.62 | 5.10 | 7.13 |
| 10 | #200 | 0.075 | | TCS | 2.21 | 3.97 | 6.46 | 2.23 | 3.99 | 6.49 | 2.24 | 4.01 | 6.53 |

## 3.2 Determination of asphalt content

The Marshall method was used to prepare 4-inch diameter asphalt concrete specimens and determine the asphalt cement (AC) content required to achieve a target air void (AV) of 4%. This target falls within the typical range of 3.0% to 5.0% specified by the Marshall mix design method (NCHRP Report 673, 2011)(National Academies of Sciences,

Engineering, and Medicine 2011). Aggregates for the nine different gradation characteristics were hot-mixed with AC type 60/70, with the AC content varied at five different levels by the dry mass of aggregate, resulting in 5 × 9 = 45 samples per aggregate type. Each hot mix was then compacted in a Marshall mold with 75 blows per side, as specified in the Marshall test procedure.

Table 1 lists the AC content values required to achieve an air void (AV) of 4% for each gradation characteristic and aggregate type. It is observed that the AC content at 4% air voids decreases with an increase in the CA, FAc, and FAf ratios. An increase in the FA ratios results in a decrease in air voids, leading to denser packing of the fine aggregates (Vavrik et al., 2002), thereby reducing the required AC content. These varying AC contents were used to prepare specimens for the resilient modulus test and the Hamburg wheel-tracking (HWT) test, which will be discussed next. In other words, the results of these tests are compared based on samples with a consistent AV of 4%.

### 3.3. Resilient modulus test

The resilient modulus test, performed in accordance with ASTM D4123 and AASHTO TP 31 standards, determines the resilient modulus ($M_r$) of bituminous mixtures through an indirect tensile test. The test applies a repeated load equal to 15% of the indirect tensile strength (ITS) at a temperature of 35ºC, with compressive loads for 160 cycles and a pulse of 100 milliseconds (1-second interval). A cylindrical asphalt concrete specimen is subjected to vertical diametral loading, typically requiring 50 to 200 repetitions. The modulus is calculated as the average of the last five cycles, using lateral deformation measured by a displacement transducer. $M_r$ can be determined as described in Eq. 4. The test setup is shown in Figure 2.

$$M_R = \frac{P(0.27+v)}{AHD} \quad \text{(MPa)} \tag{4}$$

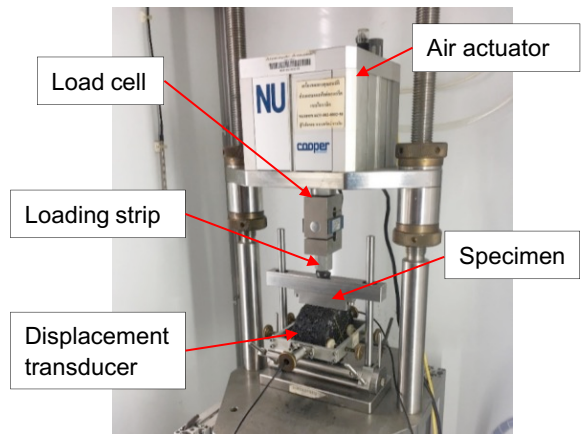

**Figure 2** Resilient modulus test setup

Table 1 also lists the average measured $M_R$ values for different Bailey's ratios and aggregate types. $M_r$ increases with higher CA, FAc, and FAf ratios. An increase in CA raises the voids in the mineral aggregate (VMA), while higher FAc and $FA_f$ ratios fill these voids, resulting in tighter packing of the fine aggregate. As these ratios increase, the mixture becomes denser and more flexible, reducing the voids and VMA (Vavrik et al., 2002).

### 3.4 Hamburg wheel-tracking test

The Hamburg wheel-tracking (HWT) test is a laboratory method used to predict asphalt performance in the field, originally developed to evaluate the rutting resistance of hot-mix asphalt (HMA). It is also suitable for assessing moisture resistance and overall stability. The test procedure follows AASHTO T 324 and is conducted under submerged conditions at 50°C. Due to its larger specimen size (6-inch), the HWT test requires preparation using the Superpave method as per AASHTO T 312. In this study, the samples were prepared with AC content to achieve an air void (AV) of 4%, as described earlier.

The Hamburg Wheel-Tracking (HWT) apparatus, widely used in Germany, measures combined rutting and moisture damage by rolling a steel wheel over an asphalt slab submerged in hot water at 50°C. The steel wheel, with a diameter of 204 mm and a width of 47 mm, generates 53±2 passes per minute. Specimens typically measure 320 mm in length, 260 mm in width, and vary from 40 to 80 mm in thickness. Cylindrical specimens have a diameter of 150 mm and a height of 60±2 mm. The device records rut depth with an accuracy of 0.01 mm and stops automatically when the preset number of passes or a rut depth of 20 mm is reached.

The HWT test produces a relationship between rut depth and the number of passes, usually showing two linear portions separated by the Stripping Inflection Point (SIP). The key parameter obtained from this test is the dynamic stability (DS), the ratio of rut depth to the number of passes at the SIP. Figures 3a and 3b show the HWT test apparatus and the HWT sample used in this study

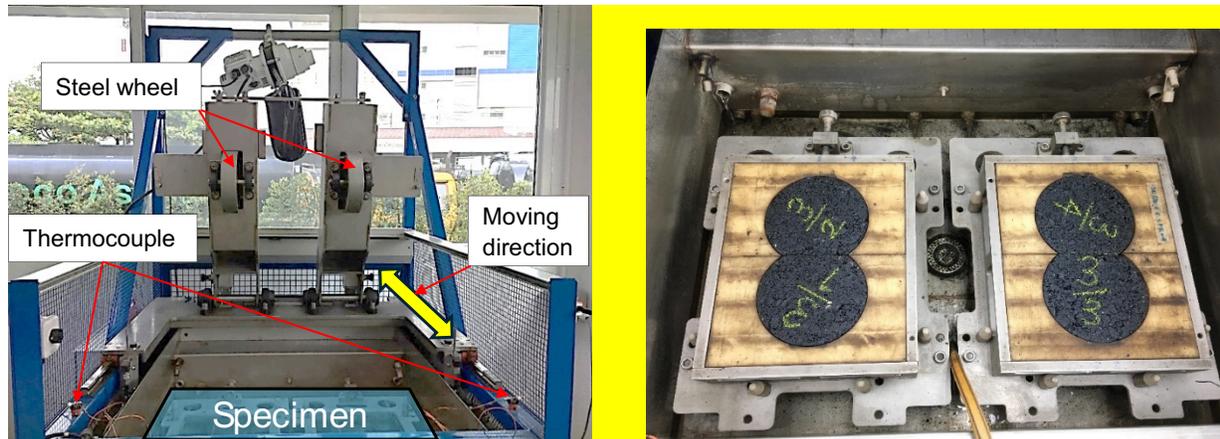

**Figure 3:** (a) HWT Test Apparatus and (b) Asphalt Concrete Sample Installed on the HWT Test Apparatus.

Table 2 also lists the average measured DS values for different Bailey ratios and aggregate types. For limestone, DS increases with higher CA, $FA_c$, and $FA_f$ ratios. Higher CA raises air voids and VMA, while $FA_c$ and $FA_f$ ratios fill these voids, leading to tighter packing of the fine aggregate and reducing VMA. Both CA and FA ratios contribute to a strong rut-resistant skeleton and adequate VMA for durability (Vavrik et al., 2002). For basalt, DS increases with higher CA ratio but decreases with higher $FA_c$ and $FA_f$ ratios. An increase in the CA ratio leads to a higher SIP, while increased $FA_c$ and $FA_f$ ratios reduce SIP resistance.

## 4. Deep learning model

This research focuses on developing an advanced predictive model for key properties of asphaltic concrete, namely resilience modulus (MR) and dynamic stability (DS), which are crucial for pavement performance. The model integrates complex input data from aggregate gradation analysis and Bailey's method, capturing the nuanced relationships between particle size distributions and material behavior. By employing explainable deep learning techniques, the study bridges the gap between black-box machine learning models and interpretable engineering insights. This approach not only predicts material properties but also elucidates the underlying mechanisms driving these predictions, offering a deeper understanding of how aggregate composition influences asphalt performance. The implementation of localized interpretability further enhances the model's utility, allowing for precise, case-specific explanations of predictions. This comprehensive methodology represents a significant advancement in asphalt mixture design, potentially revolutionizing how engineers approach the optimization of road materials. By providing both predictive power and explanatory depth, the model offers a powerful tool for creating more durable, efficient, and tailored asphalt mixtures, which could lead to substantial improvements in road infrastructure quality and longevity while potentially reducing maintenance costs and environmental impact.

## 4.1 Model architecture

This study employs a deep learning model with a diamond-shaped multi-layer perceptron (MLP) architecture, as illustrated in Fig. 4. Implemented in Python using the PyTorch library(Paszke *et al.* 2019), the network expands from 14 input features through successive hidden layers of 200, 1000, 200, 20, and 5 neurons, before converging to a single output neuron. This expansion-contraction structure enhances feature extraction capabilities. Each neuron connects to the subsequent layer via a weight matrix and bias term, following the equation:

$$y = Wx + b \qquad (5)$$

We apply the Leaky Rectified Linear Unit (LeakyReLU (Xu *et al.* 2015)) function as the activation function following each perceptron layer, defined as:

$$f(x) = \begin{cases} x \ if \ x > 0 \\ 0.01x \ if \ x < 10 \end{cases} \qquad (6)$$

LeakyReLU is chosen for several reasons: it addresses the "dying ReLU" problem by allowing a small, non-zero gradient for inactive units; it introduces necessary non-linearity; it mitigates the vanishing gradient problem during backpropagation; and it promotes sparse activation for efficient feature representation. These properties facilitate more effective training of our deep network and enable the model to learn complex patterns in the data. For the final layer, we implement a Sigmoid activation function, defined as:

$$\sigma(x) = \frac{1}{1+e^{-x}} \qquad (7)$$

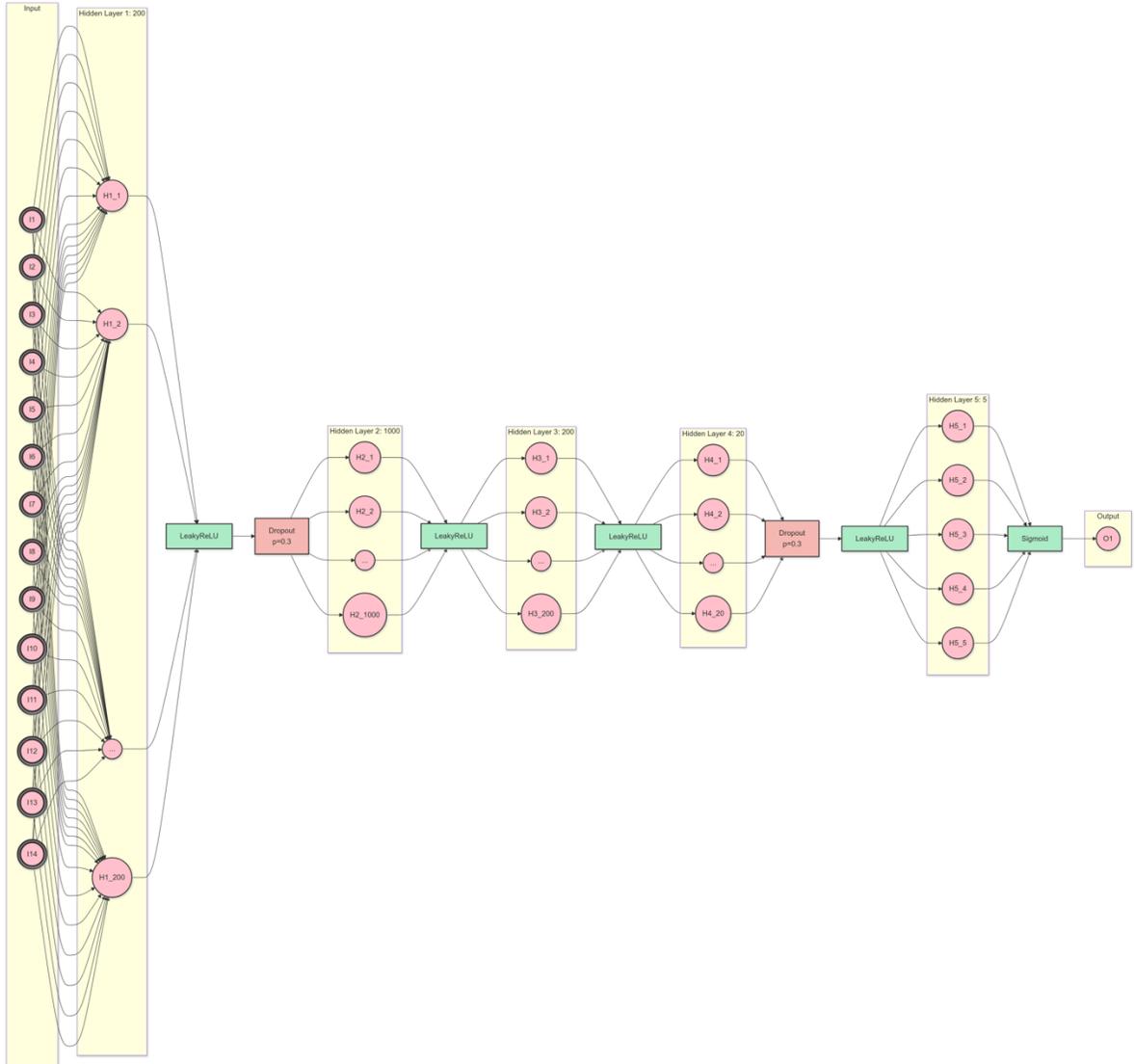

**Figure 4** The architecture of the deep learning model

## 4.2 Experiment

Firstly, the input features were normalized using a standard scaler technique. This preprocessing step is crucial for ensuring that all features contribute equally to the model and for improving the convergence of many machine learning algorithms. The standard scaler normalizes the features by subtracting the mean and dividing by the standard deviation, effectively transforming the data to have a mean of 0 and a standard deviation of 1. The equation for this normalization process is as follows:

$$X_{nor} = \frac{X - \mu}{\sigma} \tag{8}$$

Table 2 presents the model details used for training. The Adam (Adaptive Moment Estimation) (Kingma and Ba 2015) optimization algorithm was employed for gradient backpropagation to update the neural network weights. Adam combines the advantages of two other extensions of stochastic gradient descent – AdaGrad and RMSProp – by using adaptive learning rates and momentum. It efficiently handles noisy gradients and is well-suited for problems with large datasets or parameters. During the training of g, the mean squared error (MSE) was used as the loss function to compare predicted values with actual values. The MSE is calculated as follows:

$$\mathcal{L} = \frac{1}{n} \sum_{i=1}^{n} (y_i - \hat{y}_i)^2 \qquad (9)$$

This choice is motivated by several factors: it normalizes the output to a range suitable for predicting probabilities or normalized engineering property values; it enhances interpretability and prevents unrealistic predictions; its smooth gradient facilitates stable training and convergence; and its non-linear nature allows the model to capture complex relationships in the final layer. This combination of LeakyReLU in hidden layers and Sigmoid in the output layer enables our model to learn intricate non-linear relationships while producing well-bounded, interpretable predictions for our specific task of estimating asphaltic concrete properties. L2 regularization was applied during training of model, also known as Ridge regression or weight decay, is a widely-used technique in machine learning to mitigate overfitting by adding a penalty term to the model's loss function. This penalty, proportional to the square of the magnitude of the model's parameters, is mathematically expressed as:

$$\mathcal{L}(\theta) = \mathcal{L}_0(\theta) + \lambda \parallel \theta \parallel^2 \qquad (10)$$

Table 2 Model detail

| Model detail | |
|---|---|
| Loss function | Mean square error (MSE) |
| Optimization | ADAM |
| Regularization | L2 |
| Performance matrix | Mean absolute percentage error (MAPE) |

K-fold cross-validation (Jung and Hu 2015) was implemented to address potential biases in dataset partitioning and assess model generalization capability, given the limited sample size (n = 27) (Fig. 5). The dataset was partitioned into 10 equal folds. The validation process underwent 10 iterations, with each iteration designating a unique fold as the test set and the remaining 9 folds as the training set. This approach ensured that each data point was utilized once for testing and 9 times for training. Model training and evaluation occurred in each iteration, allowing for assessment of performance across various data subsets. Upon completion of all iterations, results were aggregated to provide a comprehensive evaluation of model performance. This method is particularly effective in detecting and mitigating overfitting by testing model consistency across diverse data combinations, thereby yielding a more robust assessment of predictive capabilities. The primary objective was to verify appropriate calibration of model complexity to the limited training data, thus minimizing overfitting risk. The number of folds (k) was set to 10, with 10 iterations to enhance assessment robustness. For instance, in the initial iteration, fold 1 served as the test set, while folds 2-10 were utilized for training. This systematic rotation ensured that each fold functioned as the test set exactly once across all iterations, facilitating a thorough evaluation of model performance across varied data subsets. The model's efficacy was quantified using the Mean Absolute Percentage Error (MAPE), calculated as follows:

$$\text{MAPE} = \frac{1}{n} \sum_{i=1}^{n} \left| \frac{y_i - \hat{y}_i}{y_i} \right| \times 100 \qquad (11)$$

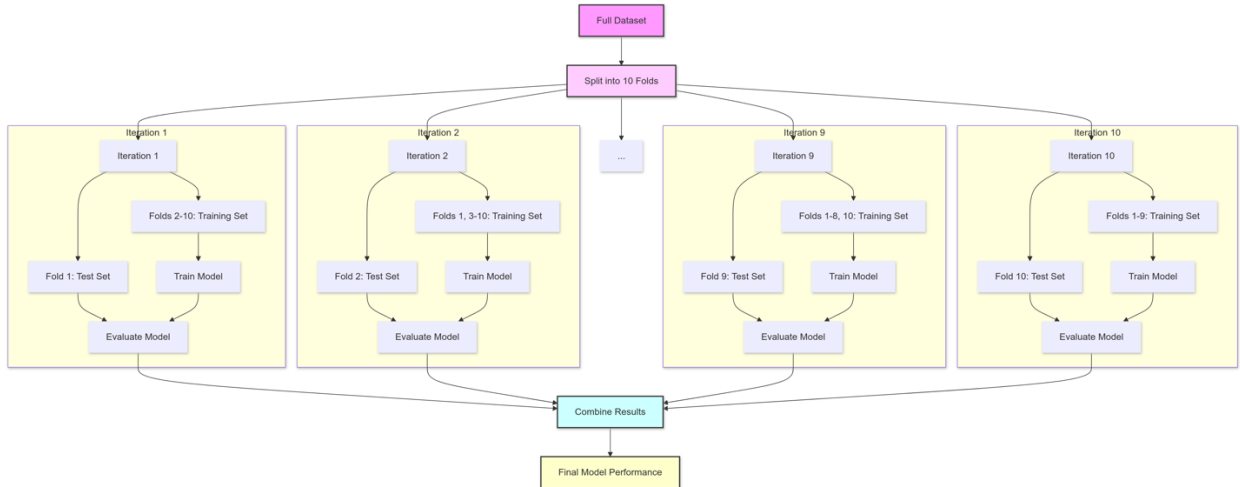

**Figure 5** The diagram for k-fold validation

The proposed model demonstrated superior performance among multilayer perceptron (MLP) architectures when compared to alternative configurations (Table 3). Increasing the number of perceptron layers resulted in lower mean absolute percentage error (MAPE) values, potentially due to overfitting. This phenomenon can be attributed to the model's increased capacity to capture complex patterns in the training data, sometimes at the expense of generalizability. Conversely, reducing the number of MLP layers led to increased MAPE values, likely due to underfitting and insufficient model complexity to capture the underlying data patterns. In comparison with other machine learning models, such as CatBoost, XGBoost, and Random Forest, the proposed model demonstrated superior performance. This superiority may be attributed to the MLP's ability to learn non-linear relationships and interactions between features without explicit feature engineering. Among the alternative models, Random Forest achieved the highest accuracy, possibly due to its ensemble nature and ability to mitigate overfitting through bagging and feature randomness. The performance differences observed across these models can be explained by their varying approaches to handling non-linearity, feature interactions, and the bias-variance tradeoff. The proposed MLP model's success suggests that the problem domain may benefit from deep learning approaches that can automatically extract hierarchical features from the input data.

**Table 3** The Results of MAPE for k-fold cross validation

| Model | Average MAPE(%) | |
| --- | --- | --- |
| | Resilience Modulus MR | Dynamic stability DS |
| CatBoost (n_estimator =1000) | 7.69 | 21.49 |
| XGBoost (n_estimator =1000) | 12.90 | 23.41 |
| Random Forest (n_estimator =1000) | 7.28 | 22.01 |
| (200:200:20:5:1) | 6.81 | 13.95 |
| (200:1000:2000:200:20:5:1) | 4.82 | 19.16 |
| Proposed Model (200:1000:200:20:5:1) | **4.61** | **14.15** |

The model was trained using a data split ratio of 80:20 for training and validation sets, respectively, to mitigate overfitting and ensure model generalization. While this split is common, future work could explore optimal ratios based on dataset characteristics or implement k-fold cross-validation for enhanced robustness. Predictive performance for Modulus of Rupture (MR) and Dimensional Stability (DS) is illustrated in Figures 6 and 7, which depict scatter plots of predicted versus actual values. The model demonstrated satisfactory predictive accuracy, achieving Mean Absolute Percentage Error (MAPE) values of 5.6% and 8.5% for MR and DS predictions, respectively. These performance metrics indicate the model's effectiveness in predicting both MR and DS with relatively low error rates. Overfitting was further mitigated through the application of regularization techniques and early stopping criteria. Subsequently, this trained model served as the foundation for implementing explainable AI (XAI) techniques, specifically SHAP (SHapley Additive exPlanations) values, to interpret feature importance and model predictions. exPlanations) values, to interpret feature importance and model predictions

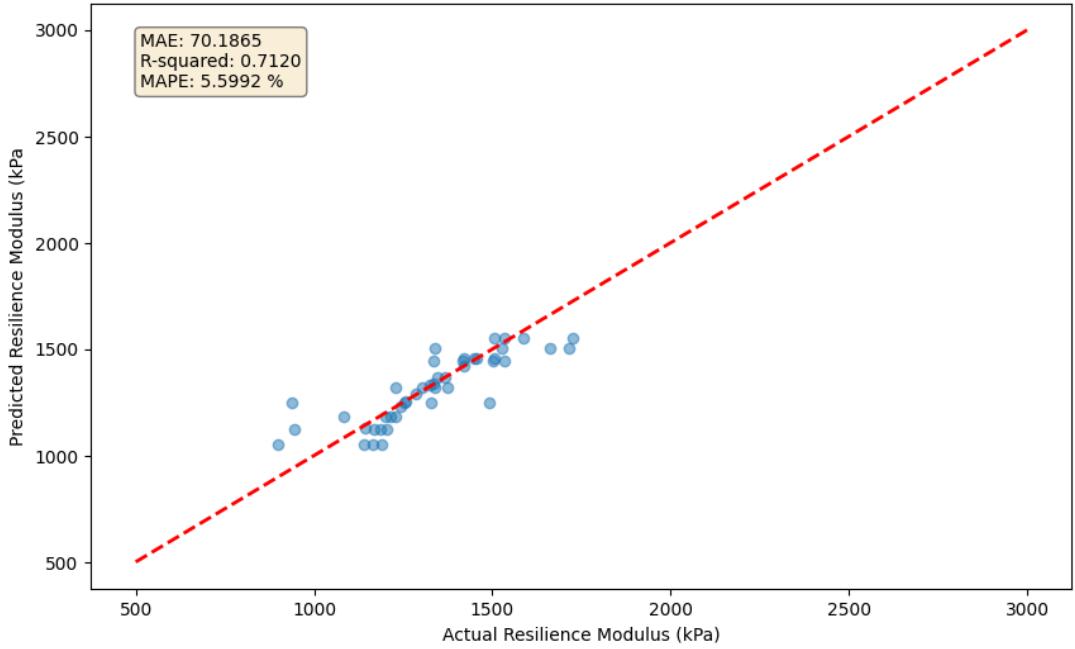

**Figure 6** The correlation between predicted resilience modulus and actual value

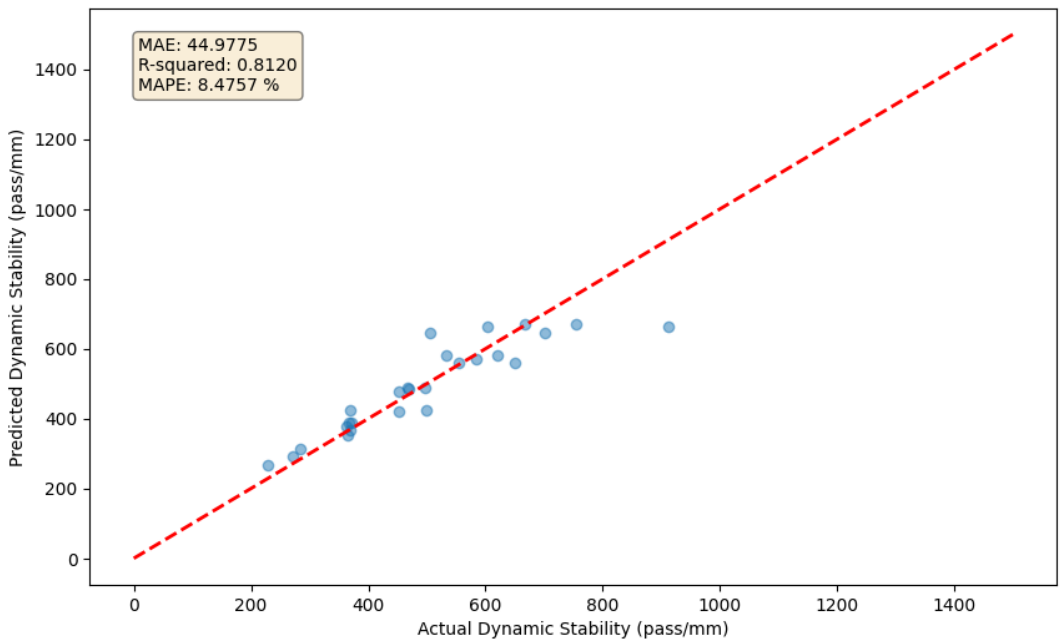

**Figure 7** The correlation between predicted Hamburg Wheel-tracking test and actual value

## 5. Explainable artificial intelligent model (XAI)

SHAP (SHapley Additive exPlanations) values, introduced by Lundberg and Lee (2017), provide a mathematically rigorous approach to interpreting machine learning model outputs. Grounded in coalitional game theory, SHAP values distribute the model's prediction f(x) among input features according to the equation

$$f(x) = \varphi_0 + \Sigma_i \ \varphi_i(x) \tag{2}$$

where $\varphi_0$ represents the base value and $\varphi_i(x)$ denotes the SHAP value for feature i. The SHAP value $\varphi_i(x)$ is computed using the Shapley value formula:

$$\varphi(val) = \sum_{S \subseteq F \setminus i} \frac{|S|!(|F|-|S|-1)!}{|F|!} [val(S \cup i) - val(S)] \tag{4}$$

This formulation considers all possible feature subsets, ensuring an unbiased attribution of feature importance. SHAP values exhibit properties including local accuracy, missingness, and consistency, enhancing their robustness for model interpretation.

The SHAP value computation process involves several steps (Fig. 8). As illustrated in the diagram, the process begins with input features (A) fed into a machine learning model (B) to generate an output (C). Concurrently, a base value (D) is established, representing the average model output across the training dataset. SHAP values (E) are then calculated using the input features (A), model output (C), and base value (D), quantifying each feature's impact on the prediction relative to the base value. The final prediction is obtained by summing all SHAP values and adding this to the base value (D), ensuring a complete explanation of the model's output. This approach provides insights into feature importance, direction of impact, interaction effects, and offers both local and global interpretability, thereby enhancing our understanding of model behavior and facilitating trust in AI systems.

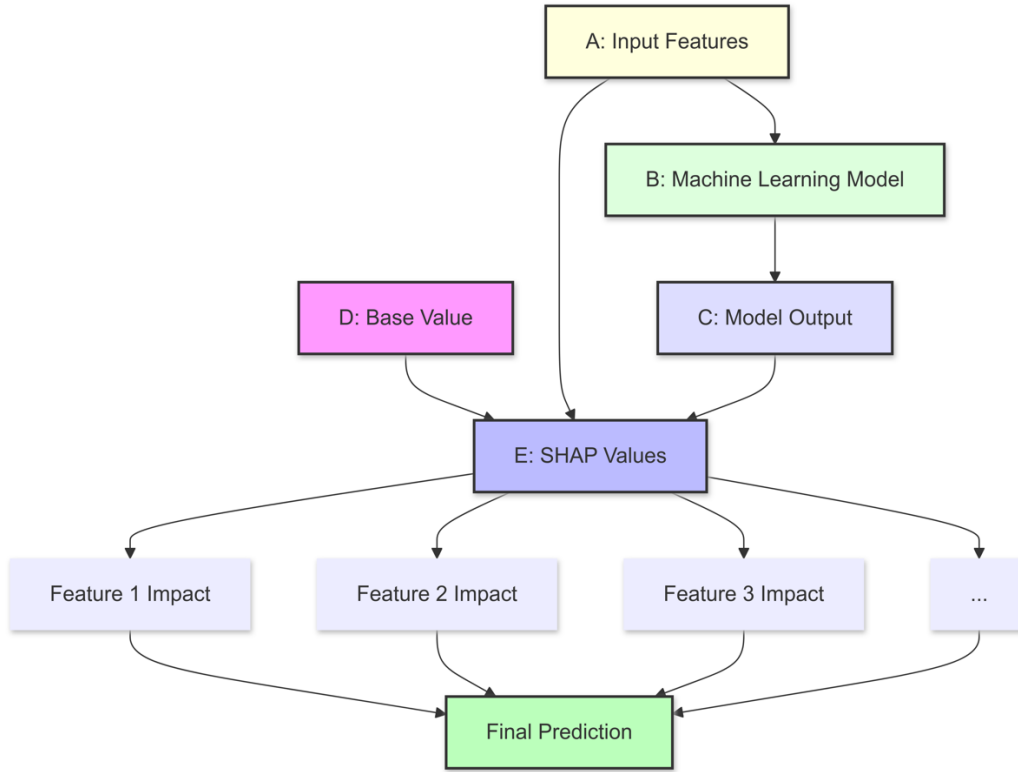

**Figure 8** The diagram for SHAP value architecture

## 5.1 Global explaination

The study utilized SHAP (SHapley Additive exPlanations) values to assess the relative importance of factors influencing Marshall Resilient Modulus (MR) and Dynamic Stability (DS) in asphalt concrete mixtures. Results revealed a size-dependent performance of aggregates, with coarse aggregates (particularly those passing through 12.5 mm and 4.75 mm sieves) significantly impacting rutting resistance (DS), while medium-fine aggregates (passing through 2.36 mm to 0.3 mm sieves) primarily influenced stiffness (MR) (Fig. 9). The 0.6 mm sieve size emerged as a critical threshold affecting both properties. Rock type demonstrated a substantial effect on DS but minimal impact on MR, suggesting its importance in rutting resistance. Fine aggregate (FA) content showed a stronger influence on MR, while coarse aggregate (CA) proportion more significantly affected DS. These findings have important implications for optimizing asphalt concrete mix designs, potentially allowing engineers to tailor mixtures for specific performance criteria by adjusting aggregate proportions and selecting appropriate rock types. The study underscores the importance of balancing fine and coarse aggregates to achieve desired stiffness and rutting resistance, while also highlighting areas for future research, such as investigating the mechanisms behind size-dependent performance and exploring optimal gradations for various conditions.

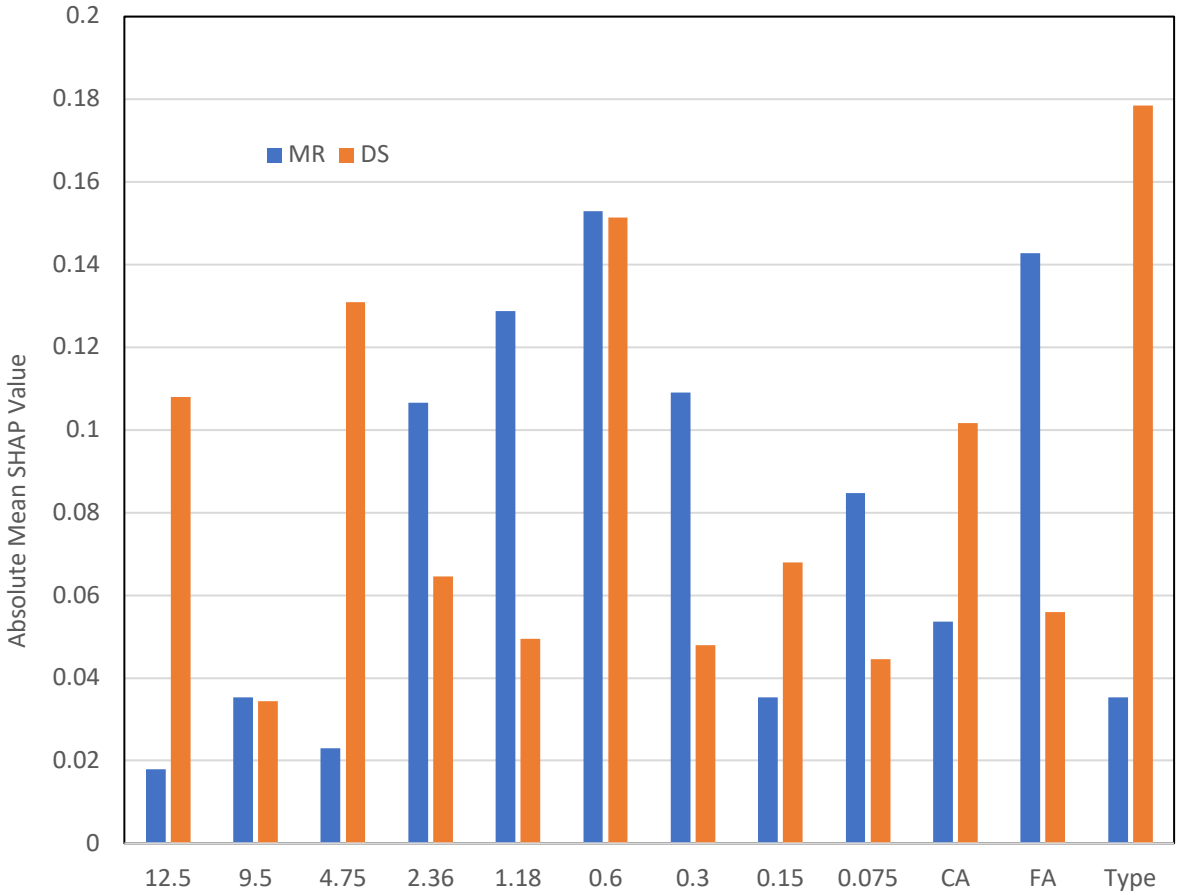

**Figure 9** The average absolute SHAP value of different feature for the resilience modulus and dynamic stability

The influence of various mix design parameters on the Dynamic Stability (DS) and Marshall Residual (MR) of asphaltic concrete was comprehensively analyzed using SHAP (SHapley Additive exPlanations) values, as illustrated in Figures 10 and 11. SHAP analysis provides a robust method for interpreting machine learning models, offering insights into the relative importance and directional impact of each feature on the target variables. The results of this analysis revealed a complex interplay of factors affecting the performance characteristics of asphaltic concrete. The percentage of particles passing through a 0.6 mm sieve emerged as the most significant factor influencing both DS and MR, exhibiting a strong inverse relationship. This finding suggests that a reduction in the proportion of particles within the 0.6-0.3 mm size range leads to a substantial increase in both DS and MR values. This relationship may be attributed to the role of these intermediate-sized particles in the aggregate skeleton, where their reduction potentially allows for better interlocking of larger particles and improved bitumen-aggregate interactions.

The Fine Aggregate Coarse Ratio (FAc Ratio), which represents the proportion of coarse particles within the fine aggregate fraction, demonstrated a positive correlation with MR, exerting a dominant influence on this parameter. This relationship may be attributed to the improved packing of aggregates within the mixture. A higher FAc Ratio indicates a greater proportion of coarser particles within the fine aggregate, which can lead to better interlocking and stability in the aggregate skeleton. This improved structure likely enhances the mixture's resistance to moisture-induced damage, as reflected in the higher MR values. Conversely, while the FAc Ratio also positively affected DS, its impact was comparatively less pronounced. This differential effect highlights the complex nature of asphalt mixture behavior under different loading conditions and performance criteria. The less significant impact on DS suggests that while the coarser portion of fine aggregates contributes to moisture resistance, other factors may play a more dominant role in determining the mixture's resistance to permanent deformation. The lithology of the aggregate used in the mixture emerged as a critical factor for DS, with the use of basalt significantly decreasing DS values. This observation underscores the importance of aggregate mineralogy and surface characteristics in determining the rutting resistance of asphaltic concrete. The lower DS values associated with basalt may be attributed to its surface texture, shape, or physico-chemical interactions with the bitumen binder.

A noteworthy observation derived from the SHAP analysis pertains to the Coarse Aggregate Ratio (CA Ratio). Specifically, a decrease in the proportion of particles retained on the 4.75 mm (No. 4) and 12.5 mm (1/2 inch) sieves, which corresponds to a lower CA Ratio, resulted in an increase in DS values. This relationship suggests that a lower proportion of these larger particles within the coarse aggregate fraction contributes to improved resistance to permanent deformation. The mechanism behind this effect could be related to a more uniform distribution of aggregate sizes, potentially leading to better particle packing and increased stability in the mixture. This improved stability likely results in better resistance to rutting under repeated loading conditions. Interestingly, the changes in the proportion of particles retained on the 4.75 mm and 12.5 mm sieves did not exhibit a statistically significant effect on MR values, indicating that the mechanisms governing rutting resistance and moisture susceptibility may differ in their sensitivity to the distribution of these particular particle sizes. This differential impact underscores the complexity of aggregate gradation and its varied effects on different performance parameters in asphalt mixtures.

This differential impact of the CA and FA Ratios on DS and MR underscores the complexity of asphalt mixture behavior and highlights the importance of carefully balancing aggregate gradation to optimize multiple performance criteria in mix design. These findings collectively emphasize the intricate relationships between mixture composition and performance characteristics in asphaltic concrete. The differential impacts of various parameters on DS and MR underscore the importance of a balanced approach in mix design, considering multiple performance criteria simultaneously. Furthermore, the results emphasize the potential for optimizing mixture performance through careful selection and proportioning of aggregate fractions, with particular attention to the critical role of intermediate-sized particles, the influence of aggregate lithology, and the optimization of both CA and FA

Ratios.

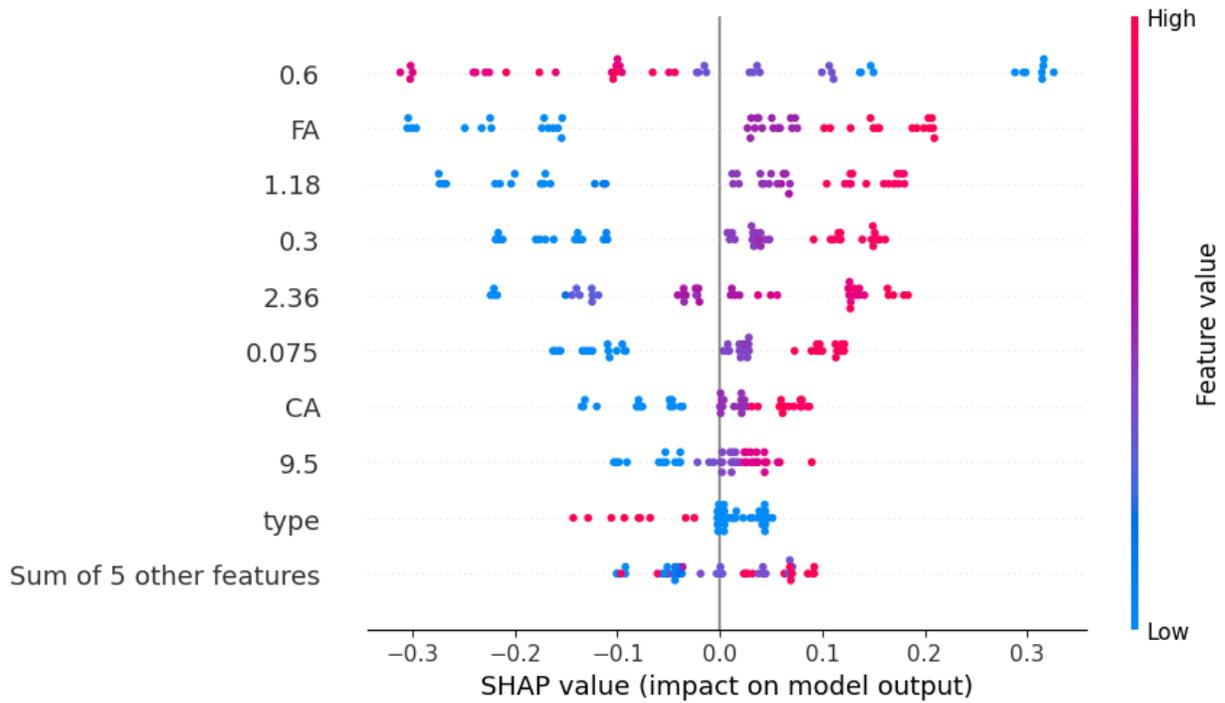

**Figure 10** The SHAP value of different feature for the resilience modulus (MR)

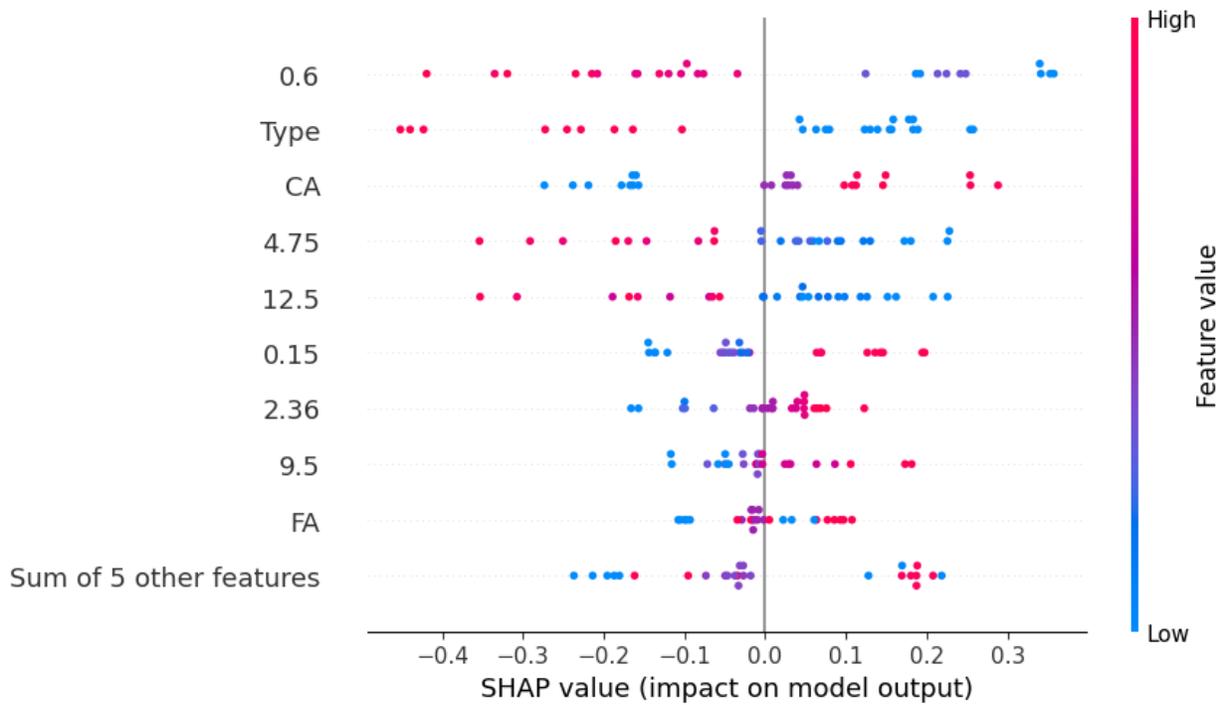

**Figure 11** The SHAP value of different feature for the Dynamic Stability (DS)

## 5.2 Local explanation

The SHAP (SHapley Additive exPlanations) values demonstrated significant utility in elucidating model predictions at a local level. This approach facilitated the decomposition of individual Marshall Residual (MR) predictions, revealing the relative contributions of each input parameter to the model's output. The analysis enabled the identification of influential parameters and provided insights into potential modifications for enhancing MR values. In a specific instance, the model predicted an MR value of 1120.381 MPa, which was below the mean value of 1196.251 MPa as shown in Fig. 12. This reduction was primarily attributed to a higher proportion of particles passing through the 0.6 mm sieve. Conversely, positive contributions to MR were associated with particles passing the 9.5 mm and 2.36 mm sieves. The coarse aggregate ratio exhibited a positive influence relative to the baseline value. To increase the MR value, the analysis suggested two potential strategies: (1) reducing the proportion of particles passing the 0.6 mm sieve, or (2) increasing the proportion of particles passing the 9.5 mm sieve. A simulation demonstrated that reducing the percentage of particles passing the 0.6 mm sieve from 14.3% to 7% resulted in a substantial increase in the predicted MR value from 1120.8 MPa to 1771.2 MPa (Fig. 13). This granular understanding of parameter impacts offers valuable guidance for engineers seeking to optimize mixture design for improved performance characteristics, illustrating the practical application of machine learning interpretability in asphalt mixture design.

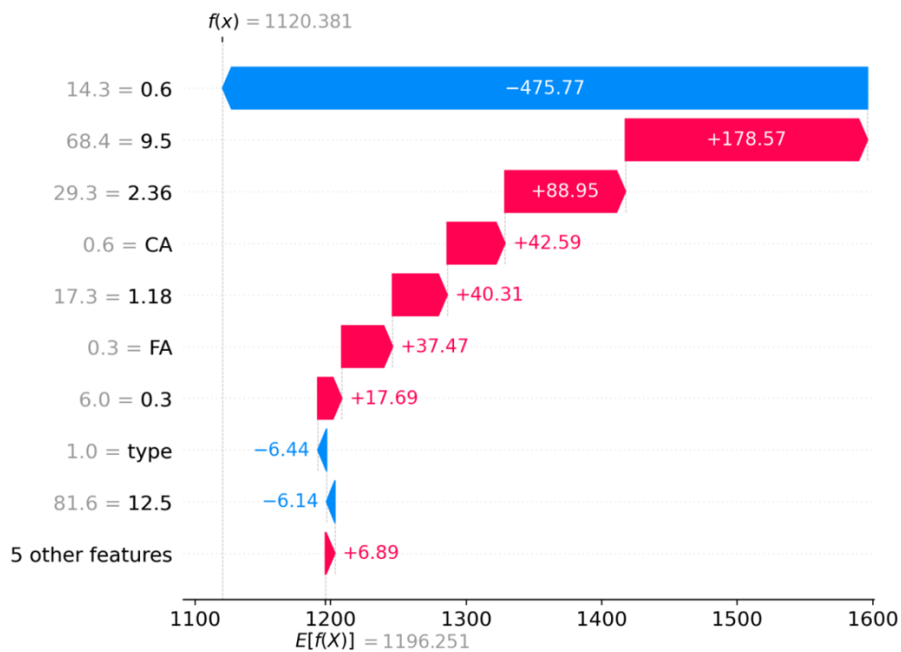

**Figure 12** The local explanation of the model for predicting value of MR

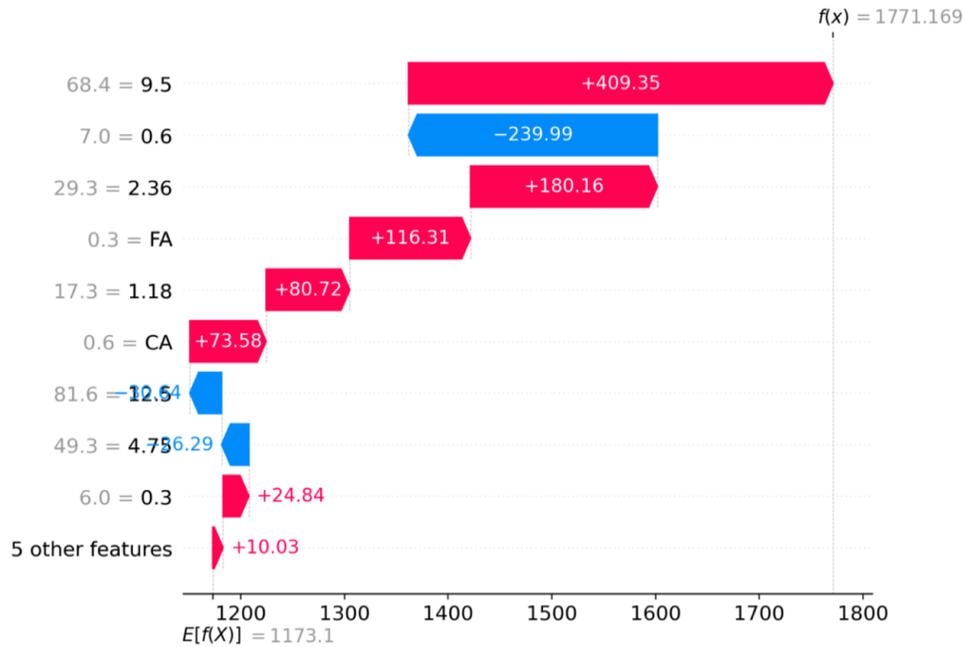

**Figure 13** The local explanation of the model for predicting value of MR after modification of gradation

The local interpretability of the model can be applied analogously to the prediction of dynamic stability (DS), as illustrated in Figs. 14 and 15. For the given sample, the model predicted a DS value of 1205 passes/mm. Analysis of feature contributions revealed that particles passing through sieve sizes of 4.75 mm, 0.6 mm, and 12.5 mm had the most significant negative impact on the DS output. Other variables demonstrated minimal effects on the DS value. To test the model's applicability for mixture optimization, we simulated a reduction in the proportion of particles passing the 4.75 mm sieve from 49.3% to 20%. This modification resulted in a substantial increase in the predicted DS to 1645.6 passes/mm. These findings underscore the dual benefits of explainable AI in this context: it not only enhances the credibility of the predicted results but also provides actionable insights for engineers to manipulate input parameters and achieve desired model outputs. This approach demonstrates the potential for AI-assisted optimization in asphalt mixture design.

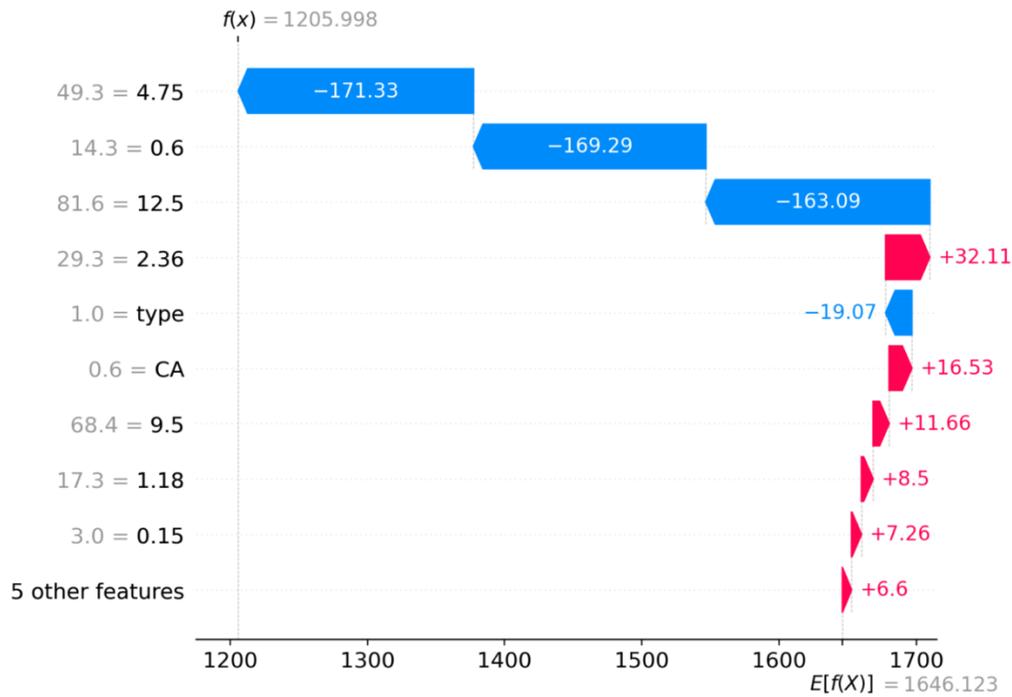

**Figure 14** The local explanation of the model for predicting value of DS

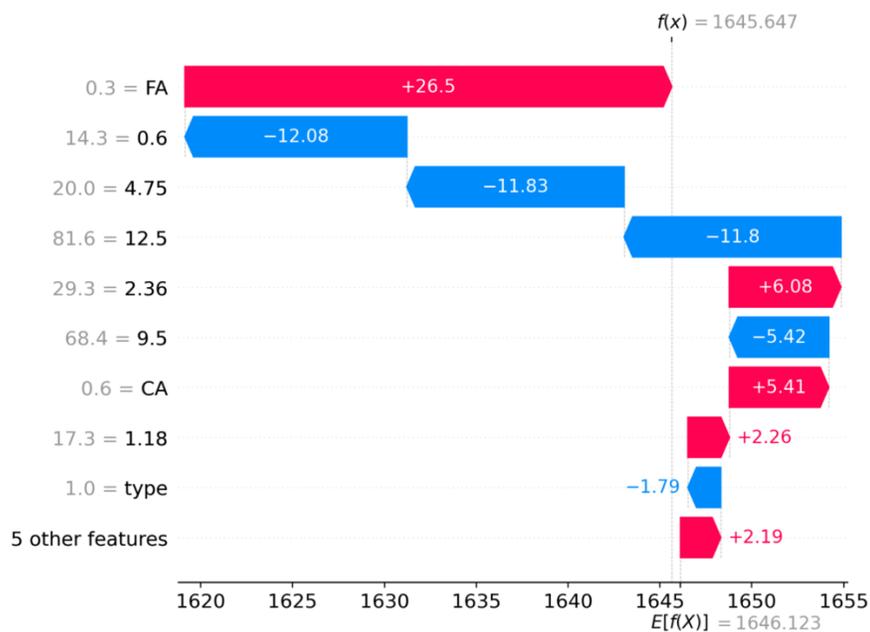

**Figure 15** The local explanation of the model for predicting value of DS after improvement

To facilitate the practical application of our predictive model, we have developed a web-based interface for estimating dynamic stability (DS) and resilience modulus (MR) of asphaltic concrete mixtures (Fig 16). The web applications are accessible via the following URLs: https://huggingface.co/spaces/Sompote/MRpredict for MR prediction and https://huggingface.co/spaces/Sompote/Dynamic.stability for DS prediction. These applications were implemented using the Streamlit library in Python, leveraging its capabilities for rapid development of data-centric web applications. The deployed versions are hosted on the Hugging Face cloud platform. This cloud-based approach ensures accessibility across various devices, including smartphones, without imposing significant computational demands on the user's hardware. The application's interface allows users to input the requisite features, specifically the aggregate gradation data and the values for coarse aggregate ratio (CA) and fine aggregate ratio (FA). Upon submission of these parameters, the application utilizes our pre-trained model to generate predictions for MR and DS. Furthermore, the interface incorporates an "Explain" function, which, when activated, provides insights into the model's decision-making process, enhancing transparency and facilitating informed interpretation of the results.

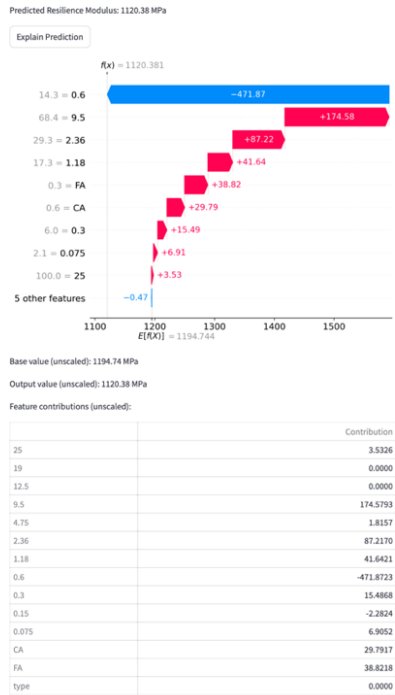

a) prediction of MR

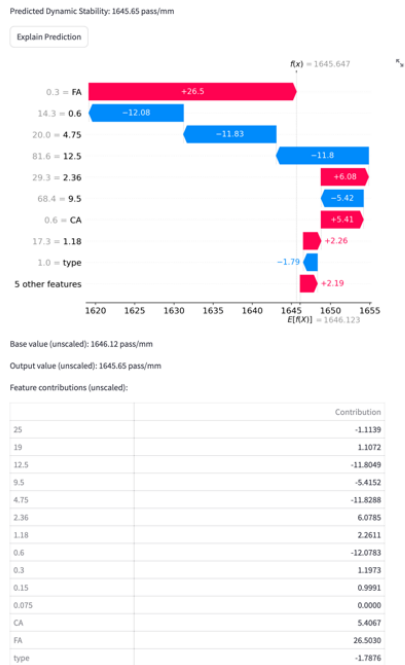

b) Prediction of DS

**Fig. 16** The web application model for using XAI model

## 6. Discussion

The findings of this study offer significant insights into the complex relationships between aggregate gradation and asphalt concrete performance, while also demonstrating the potential of explainable artificial intelligence in materials engineering. Our deep learning model's superior performance compared to traditional machine learning approaches underscores the potential of neural networks in capturing the intricate, non-linear relationships in asphalt mixture behavior. The model's ability to accurately predict both resilience modulus (MR) and dynamic stability (DS) from gradation parameters suggests that it has successfully captured key aspects of the material's internal structure and its response to loading conditions.

The SHAP value analysis revealed several crucial insights. The emergence of the 0.6 mm sieve size as a critical threshold for both MR and DS is particularly noteworthy. This finding suggests that the proportion of particles in this size range plays a pivotal role in determining the overall performance of the mixture. Future research could delve deeper into the mechanisms behind this phenomenon, potentially leading to more targeted gradation designs. The observed size-dependent performance of aggregates, with coarse aggregates primarily influencing rutting resistance and medium-fine aggregates affecting stiffness, aligns with existing theories in pavement engineering. However, our model provides a more nuanced understanding of these relationships, quantifying the relative importance of different size fractions. This knowledge could be invaluable in optimizing gradations for specific performance criteria. The substantial effect of aggregate lithology on dynamic stability highlights the importance of considering not just particle size distribution, but also the inherent properties of the aggregate material. This finding emphasizes the need for a holistic approach to mixture design, considering both gradation and material properties. The differential impacts of Coarse Aggregate (CA) and Fine Aggregate (FA) ratios on DS and MR underscore the complexity of asphalt mixture behavior. These results suggest that optimizing for one performance parameter may involve trade-offs with another, reinforcing the need for balanced design approaches.

The development of web-based interfaces for MR and DS prediction represents a significant step towards bridging the gap between research and practice. By making these tools accessible and incorporating explainable features, we're enabling practitioners to leverage advanced AI techniques in their daily work. This could potentially accelerate the adoption of more sophisticated, performance-based design methods in the industry. However, it's important to acknowledge the limitations of this study. The model's predictions are based on a specific dataset and may not generalize to all types of asphalt mixtures or environmental conditions. Future work should focus on expanding the dataset to include a wider range of materials and conditions, potentially leading to more robust and universally applicable models. Furthermore, while our explainable AI approach provides valuable insights, it's crucial to validate these findings through traditional laboratory and field testing. The correlations and relationships identified by the model should be viewed as hypotheses to be further investigated and verified.

## 7. Conclusion

This study has successfully applied explainable artificial intelligence techniques, particularly SHAP (SHapley Additive exPlanations) values, to characterize and predict the behavior of asphalt concrete with varying aggregate gradations. By focusing on two critical performance indicators - resilience modulus (MR) and dynamic stability (DS) - we have uncovered valuable insights into the complex relationships between aggregate composition and pavement performance. Our deep learning model, employing a diamond-shaped multi-layer perceptron architecture, demonstrated superior predictive capabilities compared to traditional machine learning approaches. The model's accuracy, validated through rigorous k-fold cross-validation, underscores its potential as a powerful tool for asphalt mixture design optimization.

Key findings from our SHAP value analysis reveal the critical importance of the 0.6 mm sieve size as a threshold affecting both MR and DS, size-dependent performance of aggregates, with coarse aggregates significantly impacting rutting resistance (DS) and medium-fine aggregates primarily influencing stiffness (MR), the substantial effect of aggregate lithology on DS, highlighting the importance of rock type selection in mix design, and the differential impacts of Coarse Aggregate (CA) and Fine Aggregate (FA) ratios on DS and MR, emphasizing the need for a balanced approach in gradation design. These insights provide a nuanced understanding of how various gradation parameters influence asphalt concrete performance, offering potential pathways for mix design optimization.

The development of web-based interfaces for MR and DS prediction, incorporating explainable features, further bridges the gap between research and practical application. This research contributes to the field by demonstrating the value of explainable AI in unraveling the complexities of asphalt mixture behavior. By providing both predictive power and interpretable insights, our approach offers a promising framework for creating more durable, efficient, and tailored asphalt mixtures.

## 8. Acknowledgment

The authors would like to acknowledge the support from Department of Civil Engineering, King Mongkut's University of Technology Thonburi, Thailand (CE-KMUTT 6609).

**Data Availability Statement:** The data that support the findings of this study are available from the corresponding author upon reasonable request.

**Notations**

| | |
|---|---|
| $CA$ | The coarse aggregate ratio |
| $FAf$ | The fine aggregate fine ratio |
| $FAc$ | The fine aggregate coarse ratio |
| $M_R$ | The resilience modulus |
| $P$ | The applied load, N |
| $D$ | The thickness of specimen, mm |
| $\Delta H$ | The recoverable horizontal deformation, mm |
| $\nu$ | The poisson's ratio |
| y | The output from perceptron |
| W | The weight matrix of perceptron |
| b | The bias |
| x | The input value |
| $y_i$ | The ground truth value |
| $\hat{y}_i$ | The predicted value from model |
| $\mathcal{L}(\theta)$ | The total loss |
| $\mathcal{L}_0(\theta)$ | The initial loss value |
| $\lambda$ | The coefficient for weight |
| $\theta$ | The weight of mode |
| MAPE | The mean absolute percentage error |
| $\varphi_0$ | The base SHAP value |
| $\varphi_i(x)$ | The SHAP value for feature i |
| F | The set of all features |
| Val(S) | The model prediction using feature subset S |
| $X_{nor}$ | The normalize value |
| $X$ | The input value |
| $\sigma$ | The standard deviation |
| $\mu$ | The mean value |
| $M_R$ | The resilience modulus |